\documentclass[]{article}
\usepackage{proceed2e}
\usepackage{times}
\usepackage{graphicx}
\usepackage{color}
\usepackage{amsfonts, amsmath, amsthm}
\usepackage[authoryear]{natbib}
\usepackage{hyperref}
\usepackage{color, colortbl}
\title{Influence diagrams for the optimization of a vehicle speed profile}

\author{V\'aclav Kratochv\'{\i}l \ and \ Ji\v{r}\'{\i} Vomlel\\
Institute of Information Theory and Automation\\
Czech Academy of Sciences\\
Pod vod\'{a}renskou v\v{e}\v{z}\'{\i} 4, Prague, 182 08, Czech Republic}

\theoremstyle{definition}
\newtheorem{definition}{Definition}
\newtheorem{example}{Example}

\theoremstyle{remark}
\newtheorem*{remark}{Remark}

\DeclareGraphicsExtensions{.jpg,.png,.pdf,.pdftex}
\begin{document}

\maketitle

\begin{abstract}
Influence diagrams are decision theoretic extensions of Bayesian networks.
They are applied to diverse decision problems.
In this paper we apply influence diagrams
to the optimization of a vehicle speed profile.
We present results of computational experiments in which an
influence diagram was used to optimize the speed profile
of a Formula~1 race car at the Silverstone F1 circuit.
The computed lap time and speed profiles correspond well to those achieved
by test pilots. An extended version of our model that considers a more
complex optimization function and diverse traffic constraints
is currently being tested onboard a testing car by a major car manufacturer.
This paper opens doors for new applications of influence diagrams.
\end{abstract}

\section{INTRODUCTION}

Optimization of a vehicle speed profile is a well known problem studied
in literature. Some authors minimize the energy
consumption~\citep{monastyrsky-golownykh-1993, chang-morlok-2005, saboohi-farzaneh-2009, hellstrom-2010, mensing-2011, rakha-2012}
while others aim at minimizing the total time~\citep{velenis-tsiotras-2008}.

In this paper we describe an application of influence diagrams
to the problem of the optimization of a vehicle speed profile.
Speed profile specifies the vehicle speed at each point of the path. 
We illustrate the proposed method using an example of the speed profile optimization
of a Formula~1 race car at the Silverstone F1 circuit~\citep{velenis-tsiotras-2008}.
The goal is to minimize the total lap time.
This example will be used throughout the paper
to explain the key concepts and for the final experimental evaluation of the proposed approach.
An advantage is that the optimal solution is known~\citep{velenis-tsiotras-2008}.
This allows us to compare the influence diagram solution with the analytic one.
Both solutions have a close correspondence.

The proposed method allows applications of influence
diagrams to more complex scenarios of a speed profile optimization.
Speed constraints can be invoked not only by path radii,
but also by other causes like traffic regulations, weather conditions, distance
to other vehicles, etc. Moreover, these conditions can be changing dynamically.
Also, the criteria to be optimized need not be the total time only. 
We can consider also safety, fuel consumption, etc.
We believe that influence diagrams are very appropriate for these situations since
optimum policies are precomputed for any speed the vehicle can attain. 
The optimal speed profile can be quickly updated if the conditions change.

There are two key properties that allow efficient computations.
The first one is that the overall utility function
is the sum of local utilities in all considered segments of the vehicle path.
This is the case not only when the goal is to minimize the total time,
but also when we aim at the minimal total fuel consumption
or a linear combination of these two.
The second key property is the Markov property. 
This allows to aggregate the whole future in one probability and one utility potential.  
These potentials are defined over the speed variable in the current path segment.

The paper is organized as follows. In Section~\ref{sec-physics}, we
describe the physical model of a vehicle and define the problem of the vehicle speed profile optimization.
In Section~\ref{sec-id}, we introduce influence diagrams and
in Section~\ref{sec-id-speed-profile} we apply them to the vehicle speed profile optimization.
The results of numerical experiments with real data are presented
in Section~\ref{sec-experiments}. Section~\ref{sec-related-work} reviews the related work. 
In Section~\ref{sec-conlusions}, we conclude the paper 
by a summary of our contribution and by a discussion of our future work.

\section{PHYSICAL MODEL OF THE VEHICLE\label{sec-physics}}

First, we describe a simple physical model of a vehicle. 
The content of this section is based on~\citep{velenis-tsiotras-2008}.
Although this model is too simple to model the complex behavior of a car-like vehicle,
it is sufficient for the optimization of a vehicle speed profile.

\begin{figure}[htb]
\begin{center}
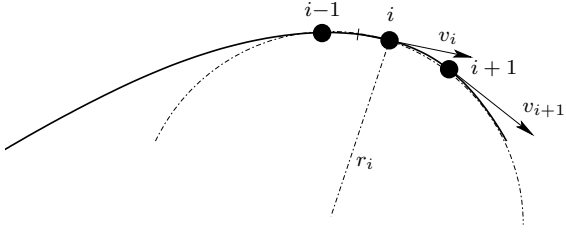
\caption{A point mass moving along a path.\label{fig-point-mass}}
\end{center}
\end{figure}

We model a vehicle as a point mass moving along a path, see Figure~\ref{fig-point-mass}.
We split the path into $n \in \mathbb{N}$ small segments of a specified length
(e.g., $5$ meters).
Let $s$ denote the length of each segment, $i \in \{0,\ldots,n\}$ be the path length coordinate,
$[i,i+1]$ be the segment between path length coordinates $i$ and $i+1$.
We assume that the acceleration is constant at each segment.
Let $v_i$ be the velocity at $i$,
and $a_i$ the acceleration at the segment $[i,i+1]$.
The velocity $v_{i+1}$ at $i+1$ is a function of the velocity $v_i$ and acceleration $a_i$:
\begin{eqnarray}
v_{i+1} & = & v_{i+1}(v_i,a_i,s) \ \ = \ \ \sqrt{(v_i)^2+2 \cdot s \cdot a_i } \label{eq-velocity}
\enspace . \rule{4mm}{0mm}
\end{eqnarray}
Time $t_{i+1}$ spent at the path segment $[i,i+1]$ is:
\begin{eqnarray}
t_{i+1} & = & t_{i+1}(v_{i}, v_{i+1}, s) \ \  =  \ \ 
s \cdot \left({\frac{v_i + v_{i+1}}{2}}\right)^{-1} \label{eq-time} . \rule{4mm}{0mm}
\end{eqnarray}
The vehicle is controlled by a control variable $u_i$, which is assumed to be constant at the segment $[i,i+1]$.
The control variable $u_i$ takes values from interval $[-1,+1]$,
where negative values represent braking and positive values
represent accelerating. We use variable $u_i$ to control acceleration $a_i$:
\begin{eqnarray}
a_i & = & \left\{  \begin{array}{ll}
          a_t^{max} \cdot u_i - c_v \cdot (v_i)^2 & \mbox{if $u_i \geq 0$}  \\
          a_t^{min} \cdot u_i - c_v \cdot (v_i)^2 & \mbox{if $u_i < 0$ \ ,}
        \end{array}
        \right. \label{eq-acceleration}
\end{eqnarray}
where $a_t^{max}$ and $a_t^{min}$ are engine and brakes characteristics, namely the maximum tangential acceleration and deceleration, respectively. $c_v$ is the deceleration coefficient for aerodynamic drag.
\begin{example}
For a F1 race car:
\begin{eqnarray}
\lefteqn{a_i \ \  = \ \  a_i(u_i,v_i)}\nonumber\\ 
& = & \left\{  \begin{array}{ll}
          16 \cdot u_i - 0.0021 \cdot (v_i)^2 & \mbox{if $u_i \geq 0$}  \\
          18 \cdot u_i - 0.0021 \cdot (v_i)^2 & \mbox{if $u_i < 0$ \ .}
        \end{array}
        \right. \label{eq-acceleration-f1}
\end{eqnarray}
\end{example}

The vehicle path is characterized by a radius profile, which is defined
as the radius $r_i$ of the circular arc which best approximates the path curve at each point $i=1,\ldots,n$
(see Figure~\ref{fig-point-mass}).
The radius $r_i$ defines the maximum speed at point $i$ as
\begin{eqnarray}
v_i & \leq & v_i^{max} \ \ = \ \ \sqrt{a_n^{max} \cdot r_i} \enspace \label{eq-vmax},
\end{eqnarray}
where $a_n^{max}$ is the maximum lateral acceleration.

\begin{example}\label{ex-2}
For a typical F1 race car $a_n^{max} = 30 \ m \cdot s^{-2}$. This implies that
\begin{eqnarray}
v_i^{max} & = &  \sqrt{30 \cdot r_i \enspace} \enspace . \label{eq-f1-vmax}
\end{eqnarray}
If $r_i=30 \ m$ then the maximum speed is $108 \ km \cdot h^{-1}$.
\end{example}
\begin{example}
In Figure~\ref{fig-radius}, we present the radius and maximum speed profiles
of the F1 Silverstone circuit (the bridge version).
The radius larger than $500$ meters is not depicted\footnote{Radius 500 meters allows maximum
speed of $441 \ km\cdot h^{-1}$ -- a speed never reached
by an F1 race car.}.
From the radius profile, we derive the maximum speed profile by use of formula~(\ref{eq-f1-vmax}).
\begin{figure}[htb]
\begin{center}
\includegraphics[width=\columnwidth]{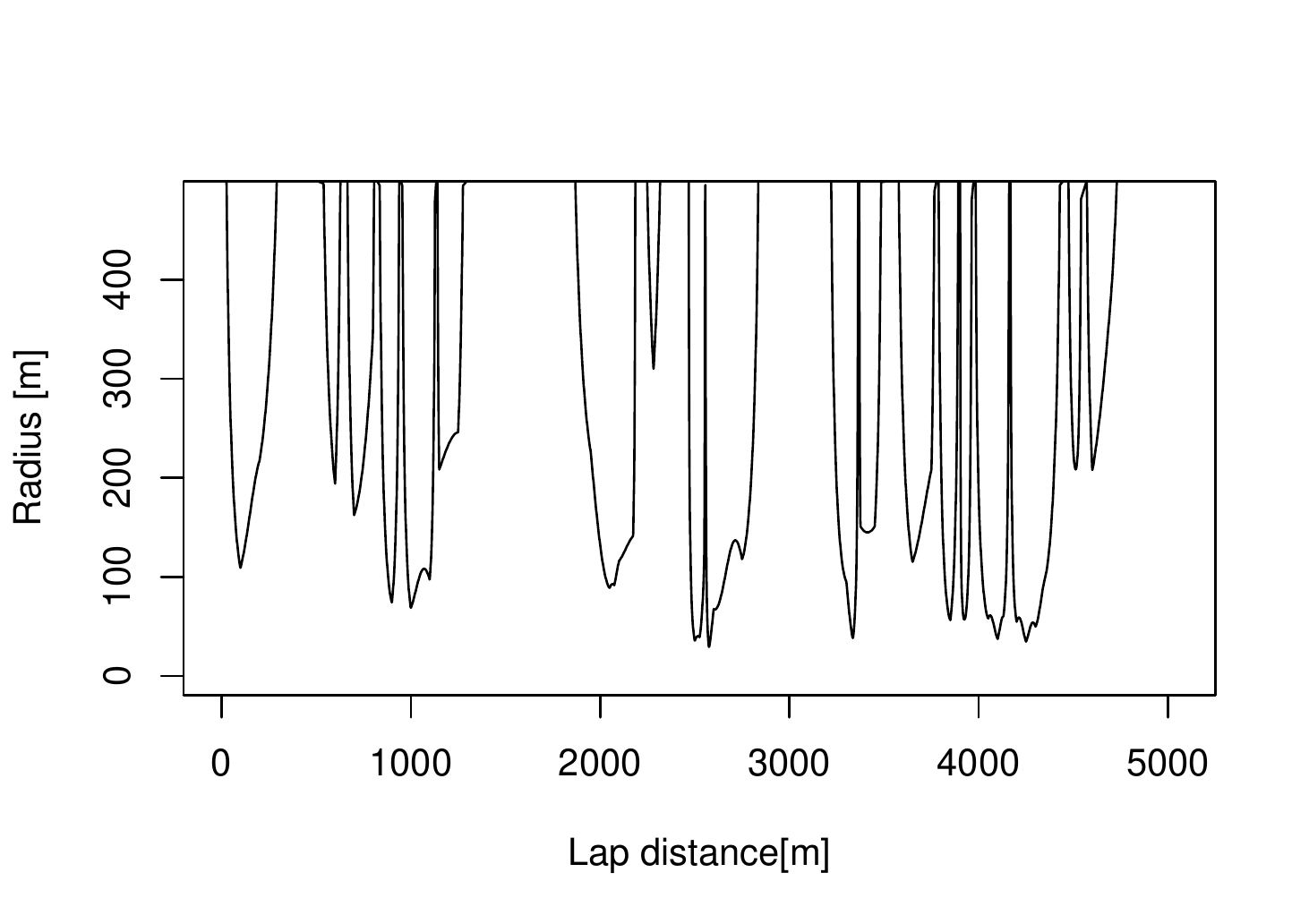}\\[-8mm]
\includegraphics[width=\columnwidth]{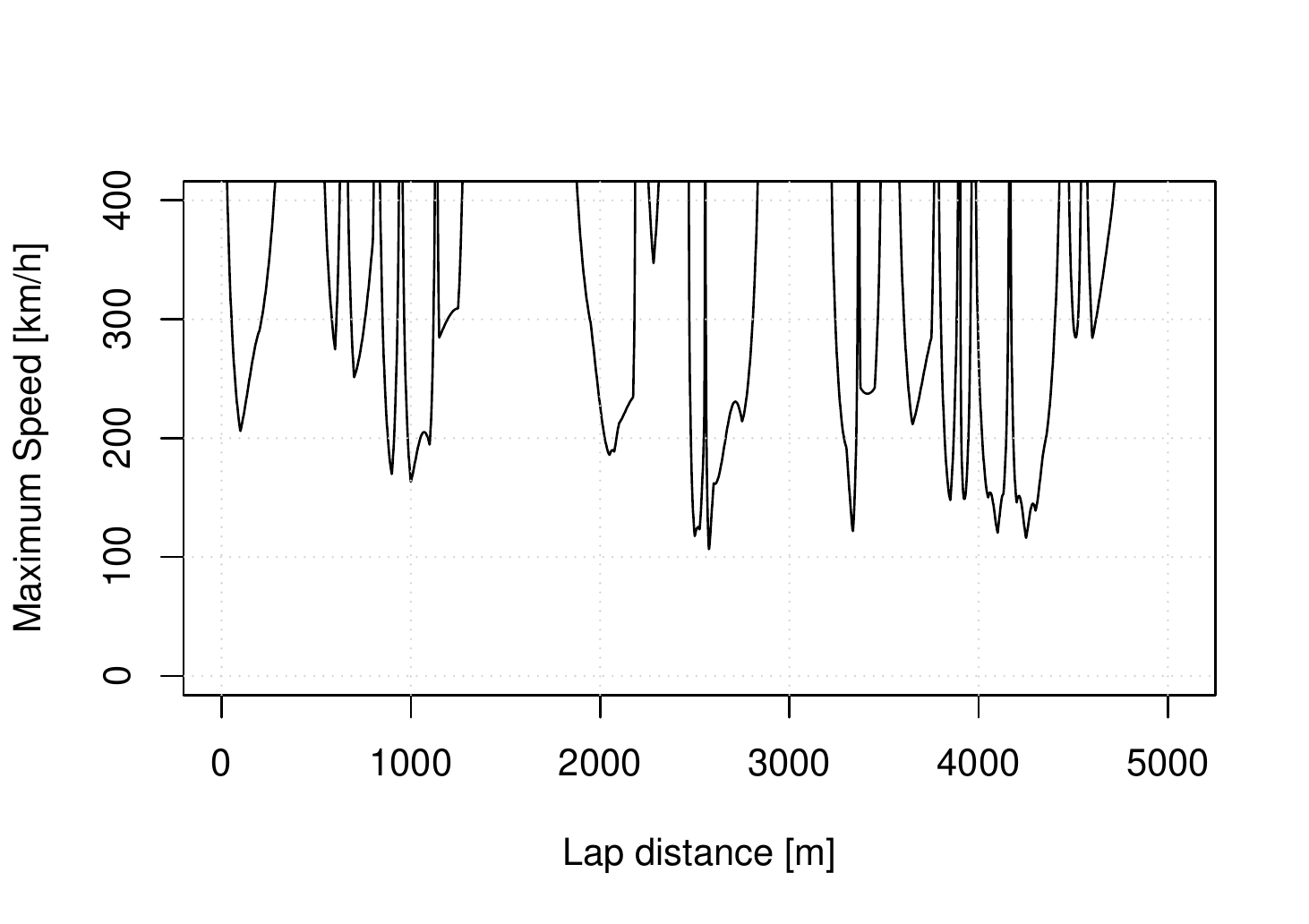}
\caption{Radius and maximum speed profiles.\label{fig-radius}}
\end{center}
\end{figure}
\end{example}

Other restrictions on maximal and minimal tangential accelerations are
due to friction forces at the tires, which 
restricts the control signals:
\begin{eqnarray}
|u_i| &  \leq & u_i^{max}(v_i) \ \ = \ \ \sqrt{1-\left(\frac{v_i}{v_i^{max}}\right)^4} \label{eq-u} \enspace .
\end{eqnarray}

Now, we can formally specify the problem.
\begin{definition}[Vehicle speed profile optimization problem]\label{def-1}
The goal is to find a vehicle speed profile $v_i, i=1,\ldots,n$ such that
\begin{itemize}
\item $v_0$ is the actual speed of the vehicle at coordinate $0$, 
\item it minimizes the total time $\sum_{i=1}^n t_i$,
\item it satisfies the speed constraints specified by formula~(\ref{eq-vmax}) for $i=0,1,\ldots, n$ and
\item it satisfies the control constraints specified by formula~\eqref{eq-u} for $i=0,1,\ldots, n$.
\end{itemize}
\end{definition}

\begin{remark}
Please, note that an optimal speed profile can be also specified 
by values of control variable $u_i$ for $i=0,\ldots,n-1$, from which it is computed.
\end{remark}

\section{INFLUENCE DIAGRAMS\label{sec-id}}
An \emph{influence diagram}~\citep{howard-matheson-1981} is a Bayesian network augmented with decision variables
and utility functions. 
In graphs, random variables are depicted as circles, decision variables as squares, and utility functions as diamonds. As an example, see Figure~\ref{fig-id}. Random and decision variables are denoted by capital letters, their states by respective lower-case ones.

A solution to the decision problem described by an influence diagram consists of a series of decision policies for the decision variables. \emph{Decision policy} for decision variable $U$ defines
for each configuration of parents of $U$ a probability distribution over the states of $U$.
\emph{Decision strategy} is a sequence of decision policies, one for every decision variable. The goal is to find an \emph{optimal} decision strategy that maximizes the expected total utility.

\subsection{SOLVING INFLUENCE DIAGRAMS}\label{sec-solving-id}

Several methods for solving influence diagrams were proposed. A simple method was published already in~\citep{howard-matheson-1981} where influence diagrams were introduced. They proposed to unfold respective influence diagram into a  decision tree and solve it using dynamic programming. Another algorithm was developed by~\citep{shachter-1986} and it foreshadowed the future graphical algorithms. Basically, one can gradually simplify respective influence diagram by successive removing nodes from its graph. When a decision node is being removed (by maximizing expected utility), the maximizing alternative is recorded as the optimal policy. Three operations to remove graph nodes were introduced; at least one of them can be used at any given time. 
Another method is to reduce an influence diagram into a Bayesian network by converting decision variables into random variables -- the solution of a specific inference
problem in this Bayesian network then corresponds to the optimal decision policy of the influence
diagram~\citep{cooper-1988}. One can also transform an influence diagram into a valuation network and solve it using variable elimination in the valuation network~\citep{shenoy-1992}.

We decided to use a method that employs a strong junction tree~\citep{jensen-1994, jensen-nielsen-2007},
which is a refinement of methods of~\cite{shenoy-1992} and~\cite{shachter-peot-1992}.
The utility nodes are eliminated first by marrying all parents of each utility node
and by including the corresponding utility potentials to cliques containing all parents of the utility node.
It has been shown that an influence diagram can be solved exactly by message passing
performed on the strong junction tree.

To every clique $C$ in the junction tree, we associate a probability potential $\Phi_C$ and a utility potential $\Psi_C$. Let $C_1$ and $C_2$ be adjacent cliques with separator $S$. To pass a message from clique $C_2$ to clique $C_1$
potentials $\Phi_{C_1}$ and $\Psi_{C_1}$ are updated as follows\footnote{Given two potentials $\Phi$ and $\Psi$, their product $\Phi * \Psi$ and the quotient $\Phi/\Psi$ are defined in the natural way, except that $0/0$ is defined to be 0 and $x/0$ for $x\neq 0$ is undefined~\citep{jensen-1994}.}~\citep{jensen-1994}:
\begin{eqnarray}
\Phi'_{C_1} & = & \Phi_{C_1} * \Phi_S, \label{eq-phi}\\
\Psi'_{C_1} & = & \Psi_{C_1} + \frac{\Psi_S}{\Phi_S} \enspace , \label{eq-psi}
\end{eqnarray}
where
\begin{eqnarray*}
\Phi_S \ = \ \underset{C_2 \setminus S}{\mathcal{M}}\Phi_{C_2}, & & 
\Psi_S \ = \ \underset{C_2 \setminus S}{\mathcal{M}} \left( \Phi_{C_2}* \Psi_{C_2} \right)
\end{eqnarray*}
and $\mathcal{M}$ is a generalized marginalization operation. The operator  $\mathcal{M}$ acts differently
for a random variable $A$ and a decision variable $U$ of a (probability or utility) potential $\Xi$:
$$
\underset{A}{\mathcal{M}} \Xi = \sum_A\Xi,~~~~~ \underset{U}{\mathcal{M}}\Xi = \max_U\Xi \enspace ,
$$
where $\sum_A$ is a shorthand for summation over all states of $A$ and $\max_U$ denotes
maximum over all states of $U$.
For a set of variables $C$, we define $\mathcal{M}_C\Xi$ as a sequence  of single-variable marginalizations. The elimination order follows the inverse order as determined by the relation $\prec$.
In case of discrete variables, the complexity of one message passing operation is $O(|C_1| + |C_2| + |S|)$, where $|C|$ denotes the number of combinations of states of variables in $C$.

Despite its similarity with the junction tree algorithm for Bayesian networks \citep{lauritzen-1988}, only the collection phase of the strong junction tree is needed to solve an influence diagram. The maximum expected utility value can be obtained by doing the remaining marginalization in the root. Above that, we can easily get the optimal decision policy for decision variables during the message passing process. For every combination of parents of a decision variable (in our case the only parent is the speed variable), it is the alternative with the maximal expected utility in the moment of message passing.

Note that this approach is highly dependent on the process of building a junction tree from an influence diagram. The size of cliques is determining the speed of the algorithm. Consequently, the junction tree algorithm is typically infeasible for solving large influence diagrams. Fortunately, this is not the case for our application.

\section{INFLUENCE DIAGRAMS FOR VEHICLE SPEED PROFILE OPTIMIZATION\label{sec-id-speed-profile}}
\begin{figure}[htb]
\begin{center}
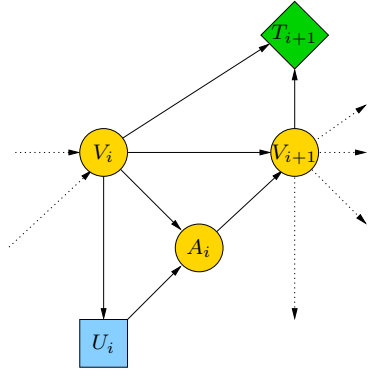
\caption{A part of influence diagram corresponding to the path segment $[i,i+1]$.\label{fig-id}}
\end{center}
\end{figure}

In this section, we use the physical model of a vehicle from  Section~\ref{sec-physics}
to create an influence diagram for the speed profile optimization.
We split the vehicle path into $n$ segments of the same length $s$.
In each segment $[i, i+1]$, $i \in \{0,\ldots, n-1\}$ there are
three random variables $V_i,A_i,$ and $V_{i+1}$, one decision variable $U_i$, 
and one utility potential $T_i$.
A part of influence diagram corresponding to one path segment is depicted 
in Figure~\ref{fig-id}. The influence diagram used in the final experiments reported 
in Section~\ref{sec-experiments} consisted of	$1010$ parts, one for a segment 5 meters long.

Variable $V_i$ corresponds to the vehicle speed in the beginning of the segment (at point $i$).
$A_i$ corresponds to the vehicle acceleration in the segment $[i, i+1]$ and
it is assumed to be constant in this segment.
$V_{i+1}$ is the vehicle speed at the end of the segment (at point $i+1$).
The decision variable $U_i$ corresponds to the vehicle control signal whose positive values denote
application of vehicle accelerator and negative values ones an application of brakes.
Utility function $T_i$ corresponds to time spent at segment $[i, i+1]$. In our implementation
the actual values of $T_i$ are \emph{time savings} achieved at segment $[i, i+1]$.
They are computed by subtracting time spent by the vehicle in the segment from a constant $t^{max}$
-- the maximum considered time\footnote{E.g., $t^{max}$ is time for
a minimal race speed $100~km/h$, which is equal to $0.036$ seconds if $s=1$ meter.}
the vehicle may spend at a segment of length $s$.
This allows us to use maximization over non-negative utility potentials, which is required
when working with random variables having states of zero probability.

In this paper we consider discrete random and discrete decision variables.
For all $i \in \{0,\ldots, n\}$, the variable $V_i$ takes its values from $\mathcal{V}$, which is a finite subset
of interval $[0,400]$ measured in $km/h$, the values of $A_i$ are from $\mathcal{A}$,
which is a finite subset of interval $[-34, 16]$ measured in $ms^{-2}$,
and values of decision variable $U_i$ are from $\mathcal{U}$, which is a finite subset
of interval $[ -100,+100]$,
where value $-100$ corresponds to the maximum braking (brakes $100\%$)
while $+100$ corresponds to full acceleration (accelerator $100\%$).
Sets $\mathcal{V}$, $\mathcal{A}$, and $\mathcal{U}$ are uniformly discretized
with discretization steps $d_V$, $d_A$, and $d_U$, respectively.
Symbols $|\mathcal{V}|$, $|\mathcal{A}|$, and $|\mathcal{U}|$ denote the cardinalities of the respective sets.

The probability and utility potentials are defined using formulas from Section~\ref{sec-physics}.
The conditional probability distributions are ``almost'' deterministic.
For each parent configuration of a variable there are only two states from the finite domain of that variable
with a non-zero probability. These two values are those that are closest to the value computed 
by the corresponding formula of the physical model of the vehicle. The conditional probability
distribution of the acceleration $A_i$ is defined as:
\begin{eqnarray}
\lefteqn{P(A_i=a|V_i=v_i,U_i=u_i) \ \ =}\nonumber\\
& & \left\{ \begin{array}{ll}
1-\frac{a_i-a}{d_A}  & \mbox{if $a = \max \{a \in \mathcal{A}, a \leq a_i\}$}\\[1mm]
1-\frac{a-a_i}{d_A}  & \mbox{if $a = \min \{a \in \mathcal{A}, a > a_i\}$}\\	
0 & \mbox{otherwise,}
\end{array}\right. \rule{3mm}{0mm}
\end{eqnarray}
where $a_i$ is defined by formula~\eqref{eq-acceleration}.

\begin{example}\label{ex-pa}
Consider $P(A_i|V_i=131km/h,U_i=25)$ and $\mathcal{A}=\{-34,-33, \ldots, -1,0,1,2,\ldots, 16\}$.
Using~\eqref{eq-acceleration-f1}
we compute the acceleration of an F1 race car $a_i = 1.2 \ ms^{-2}$.
The conditional probability distribution is specified in Table~\ref{tab-pa}.
\end{example}

\begin{table}
\begin{center}
\caption{The conditional probability table $P(A_i|V_i=131km/h,U_i=25)$}\label{tab-pa}
{\small
\begin{tabular}{|r|c|c|c|c|c|c|c|}
  \hline
  $a_i$ & \ldots & -1 & 0 & 1 & 2 & 3 & \ldots \\ \hline
  $P(a_i|v_i, u_i)$ & 0 & 0 & 0 & 0.8 & 0.2 & 0 & 0 \\
  \hline
\end{tabular}}
\end{center}
\end{table}

Similarly, we define the conditional probability distribution $P(V_{i+1}|V_i,A_i)$. 
More specifically, we combine the above approximation and formula~\eqref{eq-velocity}.
Finally, the utility function is defined as
\begin{eqnarray}
f(v_{i-1},v_i,s) & = & t^{max} - t_i(v_{i-1},v_i,s) \enspace , \rule{5mm}{0mm}
\end{eqnarray}
where function $t_i$ is defined by formula~\eqref{eq-time}.

\begin{figure}[htb]\begin{center}
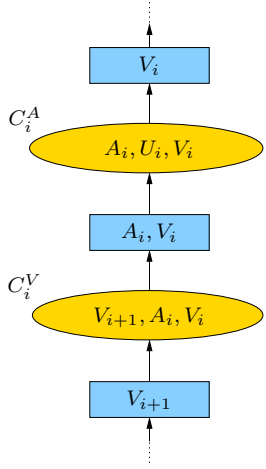
\caption{The strong junction tree for the part of influence diagram from Figure~\ref{fig-id}.\label{fig-cliques}}
\end{center}
\end{figure}

After elimination of utility nodes, the graph of the influence diagram is transformed into a strong junction tree.
See Figure~\ref{fig-cliques} where we present the strong junction tree of influence diagram from Figure~\ref{fig-id}. There are two cliques for path segment $[i,i+1]$, $i \in \{0,\ldots,n-1\}$
in the strong junction tree.
We will denote them $C_i^A$ and $C_i^V$ and define $C_i^A = \{A_i,U_i,V_i\}$
and $C_i^V = \{V_{i+1}, A_i, V_{i}\}$. 
Cliques are ordered reversely and the variable elimination is processed also in this order.
Rectangular nodes correspond to junction tree separators.

The junction tree is initialized as follows. Each conditional probability distribution and each utility function
is assigned to a clique containing all its variables. Thus
$$
\begin{array}{ll}
\Phi_{C_i^A} \ = \ P(A_i|U_i,V_i) \enspace ,&
\Phi_{C_i^V} \ = \ P(V_{i+1}|A_i,V_i)\enspace ,\\
\Psi_{C_i^A} \ = \ 0 \enspace ,&
\Psi_{C_i^V} \ = \ f(T_{i+1}|V_i,V_{i+1}) \enspace .
\end{array}
$$

\subsection{IMPLEMENTATION OF CONSTRAINTS}
During the inference we have to consider the speed and control constraints.
Both, speed and control constraints are inserted to corresponding
cliques of the junction tree. 

\subsubsection{Speed constraints}

Speed constraints are inserted in the form of likelihood evidence.
Likelihood evidence is a vector that
for each state of the corresponding variable takes values between zero and one~\citep[Section 1.4.6]{jensen-2001}.
Likelihood evidence of a speed constraint is a vector $\phi$ of length $|\mathcal{V}|$ such that
\begin{eqnarray}
\lefteqn{\phi(v) \ \ = }\nonumber \\
& & \left\{
\begin{array}{ll}
1 & \mbox{if $v \leq v_i^{max}$}\\
1-\frac{v-v_i^{max}}{d_V}  & \mbox{if $v = \min \{v \in \mathcal{V}, v > v_i^{max}\}$}\rule{5mm}{0mm}\\	
0 & \mbox{otherwise,}
\end{array}\right. \label{eq-likelihood-speed}
\end{eqnarray}
where $v_i^{max}$ is defined by~\eqref{eq-vmax}.
\begin{remark}
The idea behind the formula~\eqref{eq-likelihood-speed} is that
the closer the value of $v_i^{max}$ is to the nearest speed value $v \in \mathcal{V}$
that is greater than $v_i^{max}$ the higher is the likelihood of $v$.
In experiments, we observed that by giving a non-zero probability to the state $v$ 
just above the maximum value $v_i^{max}$ we improve the quality of results.
The coarser the discretization the larger the improvement.
\end{remark}

During the inference we include potential $\phi(V_i)$ into a clique containing $V_i$
that appears first in the computations. 
\subsubsection{Control constraints}

The control constraints~\eqref{eq-u} are applied during marginalization
of the control variable $U_i$ from a potential $\Psi_{C_i^V}'$ (we will abbreviate it as $\Xi$)
performed in the steps specified by formulas~\eqref{eq-phi} and~\eqref{eq-psi}.

For each $v_i \in \mathcal{V}$ we define a set of admissible control
\begin{eqnarray*}
\mathcal{U}'(v_i) & = & \{u_i \in \mathcal{U}, |u_i| \leq u_i^{max}(v_i)\}
\end{eqnarray*}
and compute an optimal admissible control value
\begin{eqnarray*}
u_i^*(v_i) & = & \arg\max_{u_i \in \mathcal{U}'(v_i)} \Xi(U_i=u_i,V_i=v_i) \enspace .
\end{eqnarray*}
The optimal decision policy in $U_i$ is for all $v_i \in \mathcal{V}$ 
\begin{eqnarray}
\delta_i(u|v_i) & = & \left\{\begin{array}{ll}
1 & \mbox{if $u=u_i^*(v_i)$}\\
0 & \mbox{otherwise.}
\end{array}\right.\label{eq-deterministic}
\end{eqnarray}
The value of the new potential is
\begin{eqnarray}
\Xi(V_i=v_i) & = & \Xi(U_i=u_i^*(v_i),V_i=v_i) \enspace . \label{eq-deterministic-policy}
\end{eqnarray}

However, whenever $u_i^*(v_i)$ is the least or the largest
value of $\mathcal{U}'(v_i)$ we can reduce the discretization error 
by considering also the nearest value $u_i^{**}(v_i)$ outside $\mathcal{U}'_i$.
The idea is similar to~\eqref{eq-likelihood-speed}.
If 
\begin{eqnarray*}
\Xi(U_i=u_i^{**}(v_i),V_i=v_i) & \geq & \Xi(U_i=u_i^*(v_i),V_i=v_i)
\end{eqnarray*}
then we replace the deterministic policy~\eqref{eq-deterministic} 
by a probabilistic policy
\begin{eqnarray*}
\delta_i(u|v_i) & = & \left\{\begin{array}{ll}
1-\frac{|u_i^*(v_i)-u_i^{max}(v_i)|}{d_U} & \mbox{if $u = u_i^*(v_i)$}\\
1-\frac{|u_i^{**}(v_i)-u_i^{max}(v_i)|}{d_U} & \mbox{if $u = u_i^{**}(v_i)$}\\
0 & \mbox{otherwise.}
\end{array}\right. 
\end{eqnarray*}
Formula~\eqref{eq-deterministic-policy} is replaced by
\begin{eqnarray*}
\lefteqn{\Xi(V_i=v_i) \ \ =}\nonumber\\
& & \delta_i(u_i^*|v_i) \cdot \Xi(U_i=u^*(v_i),V_i=v_i) \nonumber \\
& & + \delta_i(u_i^{**}|v_i) \cdot \Xi(U_i=u_i^{**}(v_i),V_i=v_i)
\enspace .  \label{eq-likelihood-control}
\end{eqnarray*}


\subsection{ZERO COMPRESSION}
We solve the influence method using standard strong junction tree method~\citep{jensen-1994} briefly 
described in Section~\ref{sec-solving-id}.
But the probability potentials we are working with are sparse, i.e., they contain many zeroes.
This is a consequence of conditional probability distributions being ``almost'' deterministic.
In Hugin~\citep{andersen-1990} a procedure called \emph{zero compression} is employed to improve
efficiency of inference with sparse potentials. 
In this procedure an efficient representation of the clique tables is used 
so that zeros need not be stored explicitly.
The savings can be large: in our case, we basically reduce the dimension of each table by one.
The compression does not affect the accuracy of the inference process, as it introduces
no approximations~\citep{cowell-2007}, i.e., it is an exact inference method.

\begin{example}
We can  store the distribution from Example~\ref{ex-pa} using two numbers only -- value $val$ and position $pos$ of the first non-zero number in the table. Note that the second non-zero number is positioned on $pos + 1$ with value $1-val$ and there are two non-zero numbers only. The same applies to $P(V_{i+1}|V_i,A_i)$.
\end{example}

In the standard inference method the complexity in one path segment 
is $O\left(|\mathcal{A}|\cdot |\mathcal{V}|\cdot(|\mathcal{V}| + |\mathcal{U}|)\right)$. 
In case of zero compression the complexity drops to 
$O(|\mathcal{V}|\cdot(|\mathcal{A}| + |\mathcal{U}|))$. 
In Section~\ref{sec-zero-compression} we evaluate the savings experimentally.

\section{EXPERIMENTS\label{sec-experiments}}
We performed experiments with a model of a Formula~1 race car at the Silverstone F1 circuit (the bridge version). The goal is to find a speed profile that minimizes the total lap time and satisfies speed and acceleration constraints 
as specified by Definition~\ref{def-1}.
The speed constraints are derived from radius of curves and the maximum allowed lateral acceleration $a_n^{max}$
-- see formula~\eqref{eq-vmax}. For a typical F1 race car $a_n^{max} = 30 \ ms^{-2}$ 
-- see Example~\ref{ex-2}. The acceleration constraints are defined by formula~\eqref{eq-u}.

In our experiments we use the influence diagram described in Section~\ref{sec-id}.
The experiments were conducted in the following way:
\begin{enumerate}
\item Define the length of one segment $s$ and sets of variables' states $\mathcal{V,A,U}$.
\item Initialize potentials $\Phi_{C_i^A}, \Phi_{C_i^V}, \Psi_{C_i^A},$ and $\Psi_{C_i^V}$ and set up the junction tree.
\item Insert speed and acceleration constraints to the junction tree.
\item Compute the optimal policies $\delta_i, i=0,1,\ldots,n-1$.
\item Use the optimal policy and  the initial speed\footnote{Initial speed $v_0$ 
is set as in~\citep{velenis-tsiotras-2008}.} $v_0 = 312 \ km/h$ to compute an optimal
speed profile as specified by formulas~\eqref{eq-speed-profile-1},~\eqref{eq-speed-profile-2}, and~\eqref{eq-speed-profile-3}.
\end{enumerate}
The expected speed $\hat{v}_{i+1}$ at coordinate $i+1$ is computed using 
formulas~\eqref{eq-velocity} and~\eqref{eq-acceleration-f1} from the expected control value $\hat{u}_{i}(v_i)$,
which is computed as a weighted average of policies for two values $\underline{v}_i, \overline{v}_i$ 
from $\mathcal{V}$ that are the closest to $v_i$:
\begin{eqnarray}
\hat{v}_{i+1} & = & v_{i+1}(v, s, a_{i}\left(\hat{u}_i(v_{i}),v_i)\right) \enspace ,\label{eq-speed-profile-1}\\
\hat{u}_{i}(v_i) & = & \sum_{v \in \{\underline{v}_i, \overline{v}_i\}} 
w(v,v_i) \cdot  \sum_{u \in \mathcal{U}} 
u \cdot \delta_i(u|v) \label{eq-speed-profile-2} \enspace , \rule{5mm}{0mm}\\
w(v,v_i) & = & 1-\frac{|v-v_i|}{d_V} \label{eq-speed-profile-3} \enspace . 
\end{eqnarray}

All algorithms used in our experiments were implemented 
in the programming language~R~\citep{R}.


\subsection{ZERO COMPRESSION EXPERIMENTS\label{sec-zero-compression}}

We compared computational time of the zero compression and standard junction tree inference methods,
see Table~\ref{tbl-comparison}. 
Recall that symbols $|\mathcal{V}|$, $|\mathcal{A}|$, and $|\mathcal{U}|$ denote the cardinalities of the respective sets.
The experiments were carried out on an influence diagram consisting 
of 10 path segments. We can see that zero compression 
brings large computational savings for fine grained discretizations.

\begin{table}[htb]
\centering
\caption{Comparisons of the CPU time for the zero compression and the standard approach.}
\vspace{3mm}
\label{tbl-comparison}
{\small
\begin{tabular}{|ccc|r|r|}
  \hline
$|\mathcal{V}|$ & $|\mathcal{A}|$ & $|\mathcal{U}|$ & \multicolumn{1}{c|}{zero compr. $[s]$} &  
	\multicolumn{1}{c|}{standard $[s]$} \\ 
	\hline
  50 & 50 & 50 & 0.08 & 0.73 \\
  100 & 100 & 100 & 0.20 & 4.78 \\
  100 & 100 & 200 & 0.19 & 7.09 \\

  100 & 200 & 100 & 0.25 & 9.17\\
  200 & 100 & 100 & 0.39& 13.01\\
  200 & 200 & 100 & 0.49 & 24.30\\
  200 & 200 & 200 & 0.50 & 29.61\\
  400 & 100 & 100 & 0.84 & 38.53\\
  400 & 400 & 100 & 1.26 & 142.92 \\
  \hline
\end{tabular}}
\end{table}


\subsection{DISCRETIZATION EXPERIMENTS}

We tested different variable discretizations and different path segmentations. 
The goal was to find a combination of these parameters that represents a reasonable 
tradeoff between precision and computational time. 

\begin{figure*}[htb]
\begin{center}
\begin{tabular}{cc}
\includegraphics[width=0.95\columnwidth]{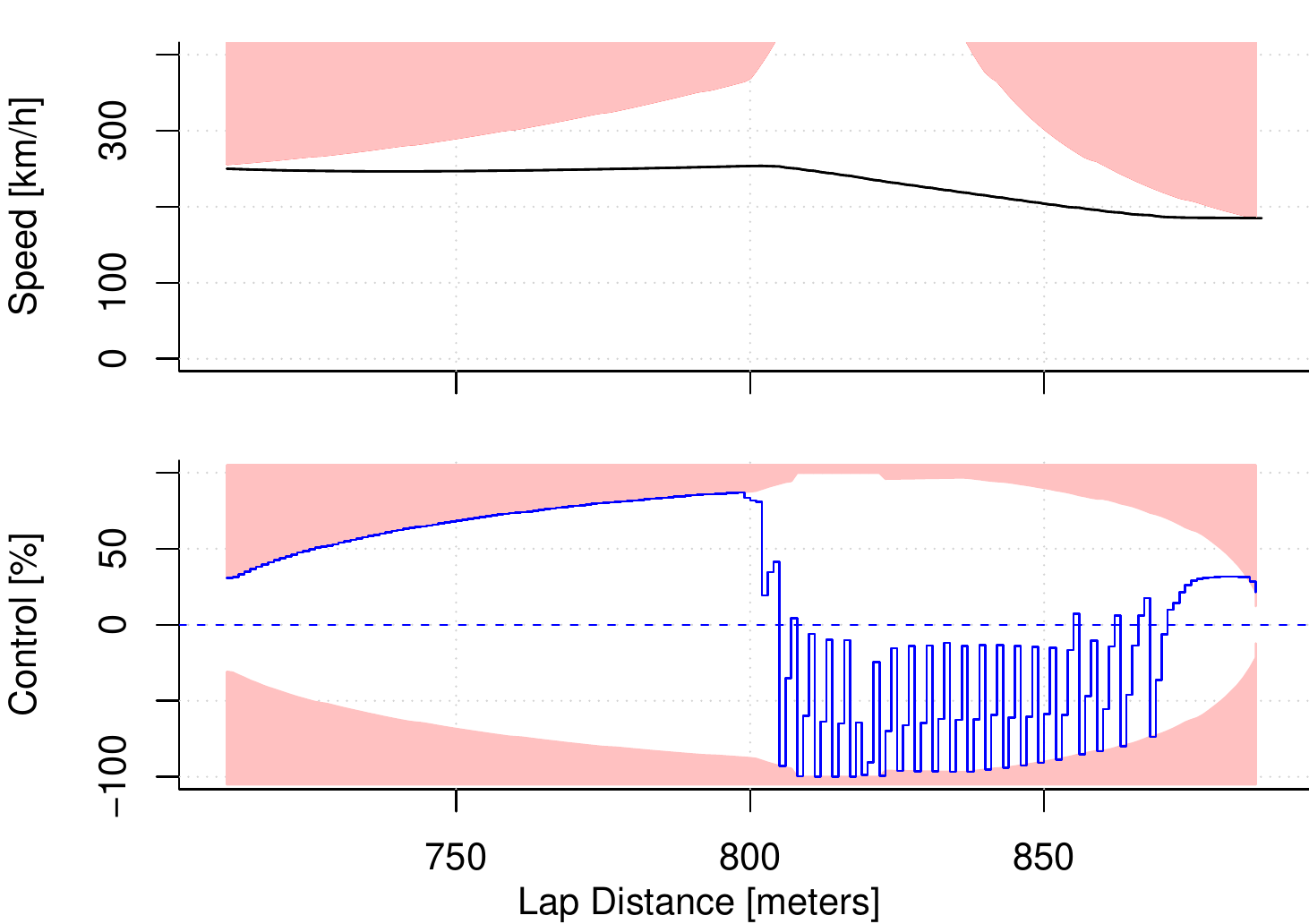} &
\includegraphics[width=0.95\columnwidth]{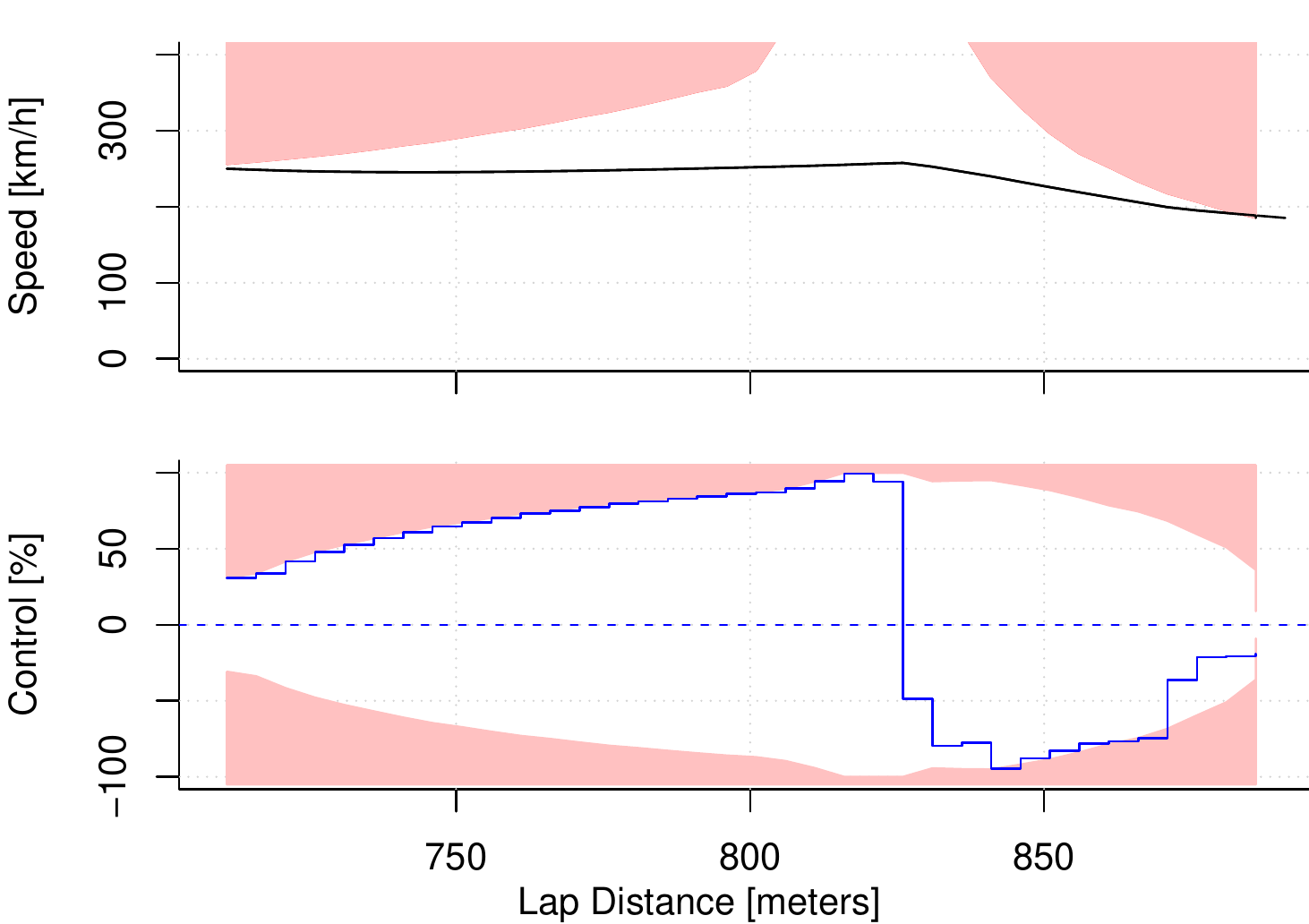}\\
Settings: $s = 1m$, $\mathcal{|V| = |A| = |U|} = 100$ &
Settings: $s = 5m$, $\mathcal{|V| = |A| = |U|} = 100$ \\[5mm]
\includegraphics[width=0.95\columnwidth]{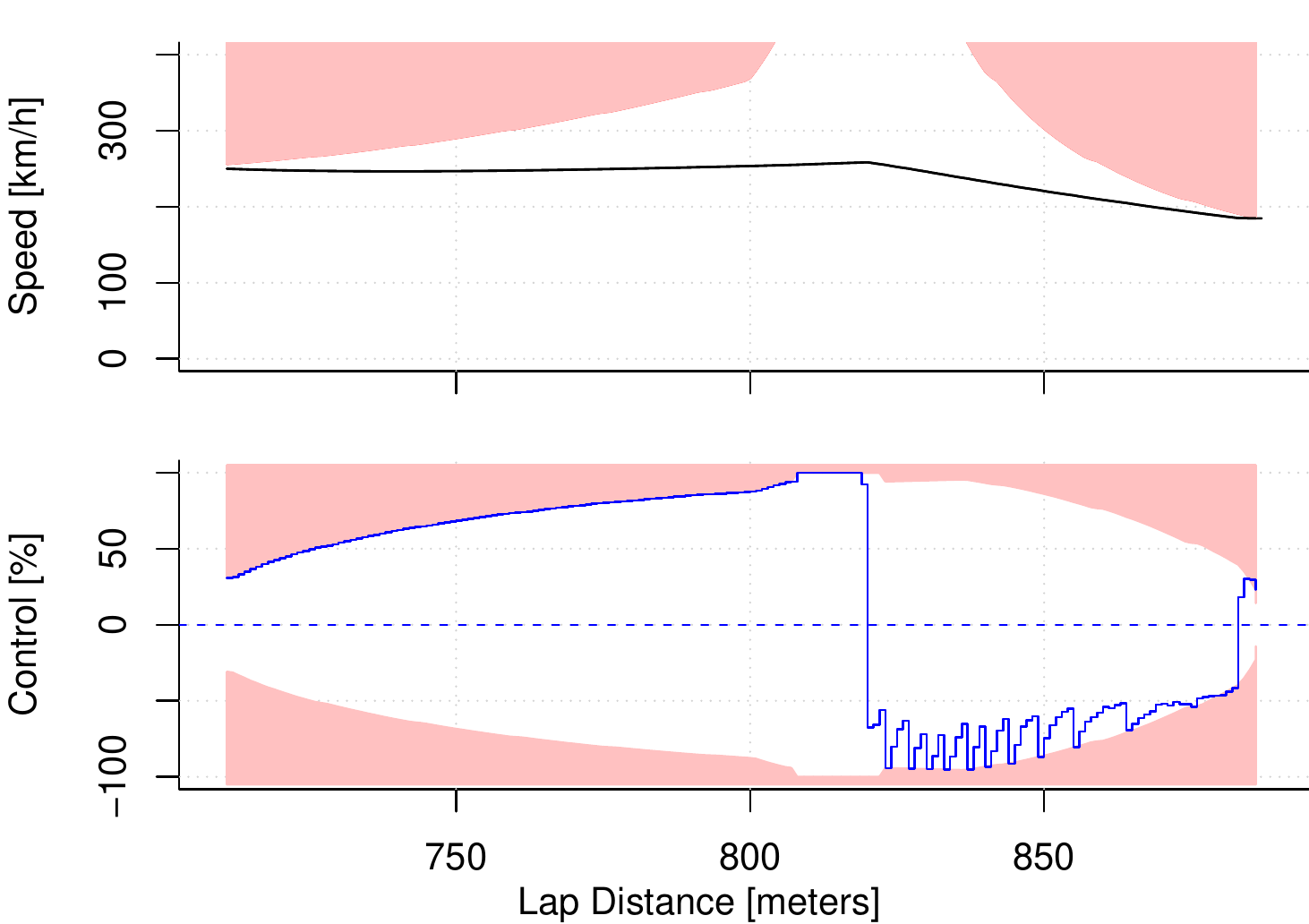} &
\includegraphics[width=0.95\columnwidth]{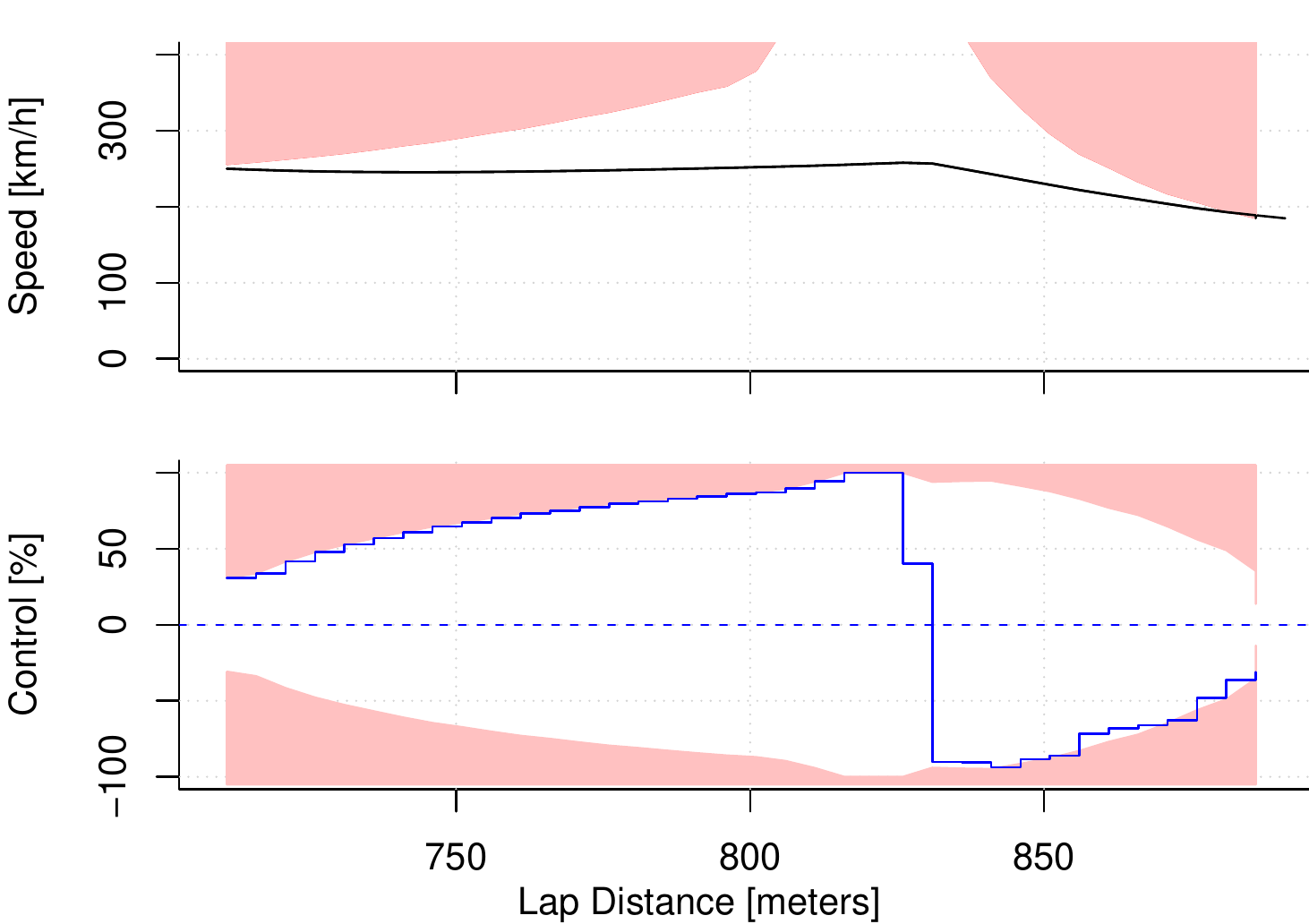} \\
Settings: $s = 1m$, $|\mathcal{V}| = 400, \mathcal{|A| =|U|} =  100$ &
Settings: $s = 5m$, $|\mathcal{V}| = 400, \mathcal{|A| = |U|} = 100$\\
\end{tabular}
\caption{Comparisons of various path segmentations and discretizations.\label{fig-discretizations}}
\end{center}
\end{figure*}

We performed experiments on a part of Silverstone F1 circuit 180 meters long.
Results of some performed experiments are presented in Figure~\ref{fig-discretizations}.
In the upper part of these figures, speed profiles are depicted while
the control profile is plotted in their lower part. 
The shaded areas are forbidden by the speed and control constraints, respectively.

It turned out that the number of states of the model variables should be in accordance with the path segmentation. 
The finer is the path segmentation, the more variables' values are required.
Discretization that is not well balanced with the path segmentation leads to oscillations of decision (control) variables 
as it is illustrated for $s = 1m$, $\mathcal{|V| = |A| = |U|} = 100$ in Figure~\ref{fig-discretizations}. 

In the experiments it turned out the oscillation depends strongly on $|\mathcal{V}|$.
Basically, there are two reasons. First, the discretization has to be able to distinguish small 
speed changes within one segment of the path. The reason can be elucidated 
by the following example.
\begin{example}
Consider uniformly accelerated motion with the initial speed $v_0 \in \{200 \ km/h, 300 \ km/h\}$. 
In case of full throttle,  the acceleration computed by formula~\eqref{eq-acceleration-f1}
is $a \doteq  9.52 \ ms^{-2}$ and $a \doteq 1.42 \ ms^{-2}$, respectively.
Considering segments of various length $s$, we can compute the speed 
at the end of the respective segment using formula~\eqref{eq-velocity}. 
From Table~\ref{tbl-speeds} we can see that in case of $s=1\ m$ the discretization has to be fine grained
in order to capture small speed changes within such a short segment.

\begin{table}[htb]
\centering
\caption{Speed [in $km/h$] in case of full acceleration 
}
\vspace{3mm}
\label{tbl-speeds}
{\small
\begin{tabular}{|c|c|c|c|}
  \hline
  $v_0$ & \multicolumn{3}{c|}{$v_1$}\\
	\cline{2-4}
	 & $s=1 \ m$ &  $s=5 \ m $  &  $s=20 \ m$  \\ 
	\hline
  200 & 200.6159 & 203.0606 & 211.9774 \\
  300 & 300.0612 & 300.3058 & 301.2215 \\
  \hline
\end{tabular}}
\end{table}
\end{example}

The second reason is the inference algorithm itself. Information 
passes through clique separators. In Figure~\ref{fig-cliques} we can see that all separators 
contain $V_i$ and every second separator consists of single $V_i$ only. 
Hence, the size of $\mathcal{V}$ limits the information flow between respective cliques. 
In other words, $|\mathcal{V}|$ represents a bottleneck of the inference mechanism. 

Representative results for the whole Silverstone F1 circuit are presented in 
Table~\ref{tbl-discretization-results}. 
The expected lap time is quite stable with respect to different discretizations.
For the final experiments we selected the configuration printed in boldface since
from those that respect the speed and acceleration constraints well it is least computationally 
demanding.

\begin{table}[htb]
\centering
\caption{CPU and lap time for diverse path segmentations and discretizations.}
\vspace{3mm}
\label{tbl-discretization-results}
{\small
\begin{tabular}{|c|c|c|c|r|r|}
\hline
$s$ &$\mathcal{|V|}$&$\mathcal{|A|}$&$\mathcal{|U|}$& lap time & CPU time \\
$[m]$ & & & &  \multicolumn{1}{c|}{$[s]$} &  \multicolumn{1}{c|}{$[s]$} \\
\hline
10 & 400 & 200 & 100 & 83.92 & 20.61 \\
10 & 800 & 400 & 200 & 83.83 & 103.94 \\
5 & 400 & 200 & 100 & 84.09 &  42.88\\
5 & 800 & 200 & 100 & 83.97 & 166.88 \\
{\bfseries 5} & {\bfseries 800} & {\bfseries 400} & {\bfseries 200} & {\bfseries 83.95} & {\bfseries 197.93} \\
5 & 800 & 800 & 800 & 83.94 & 473.16 \\
5 & 1000 & 1000 & 1000 & 83.91 & 588.80 \\
1 & 800 & 400 & 200 & 84.17& 2110.68 \\
1 & 1000 & 1000 & 1000 & 84.10 & 2850.66 \\
1 & 1600 & 1000 & 1000 & 83.96& 5544.99 \\
\hline
\end{tabular}}
\end{table}

\subsection{INFLUENCE DIAGRAM SOLUTION}

\begin{figure*}[htbp]
\begin{center}
\includegraphics[width=0.9\textwidth]{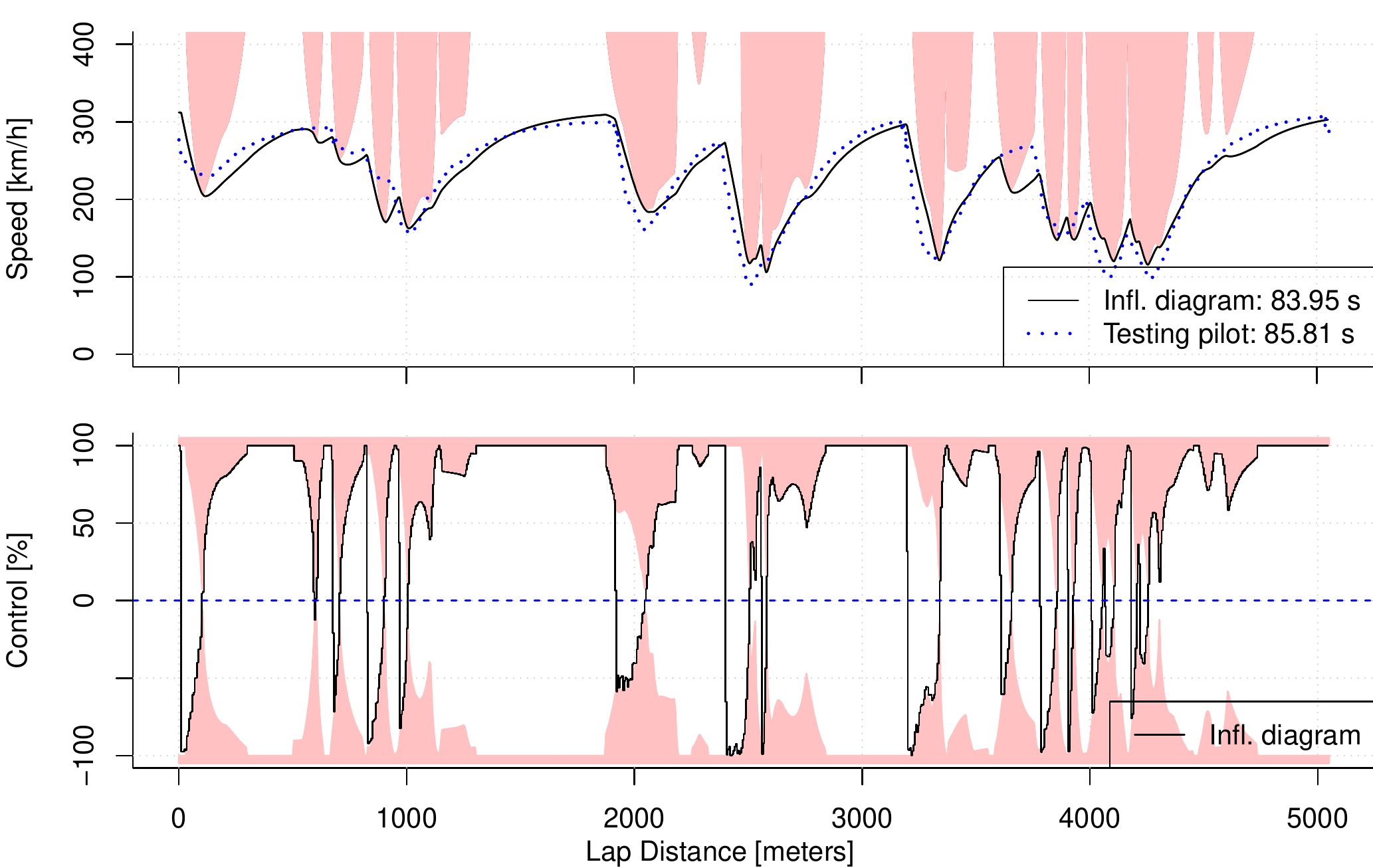}
\caption{A comparison of speed profiles: the influence diagram and 
a test pilot~\citep{gps-measurement-2002}. The shaded areas are forbidden
by the speed and control constraints, respectively.\label{fig-speed-profiles-1}}
\end{center}
\end{figure*}

We used the influence diagram to compute the speed profile for the Silverstone F1 circuit. It
is plotted in the upper part of Figure~\ref{fig-speed-profiles-1} by a full line. 
The bridge version of Silverstone circuit is $5049$ meters long, which corresponds to $1010$ segments $5 m$ long. 
In this figure we compare the computed speed profile with a test pilot performance 
at the Silverstone F1 circuit
~\citep{gps-measurement-2002}.
In the upper part of Figure~\ref{fig-speed-profiles-1} the test pilot speed profile 
is plotted by a dotted line. 
Notice that the testing pilot violates these restrictions several times.
Also, the test pilot acceleration is slower than expected. 
The speed constraints used in the model seem to be
too cautious and the car acceleration ability a bit exaggerated.

It is interesting to compare the total lap time estimated by the influence 
diagram model with results achieved by F1 pilots. While the model estimated time $83.95$ seconds
is little lower than time achieved by the test pilot -- $85.51$ seconds, it is higher
than the fastest ever lap time -- $78.12$ seconds -- attained by Sebastian Vettel with his Red Bull-Renault
in the qualification of the 2009 British Grand Prix. 

\subsection{ANALYTIC SOLUTION}\label{sec-experiments-analytic}

\begin{figure*}[htbp]
\begin{center}
\includegraphics[width=0.9\textwidth]{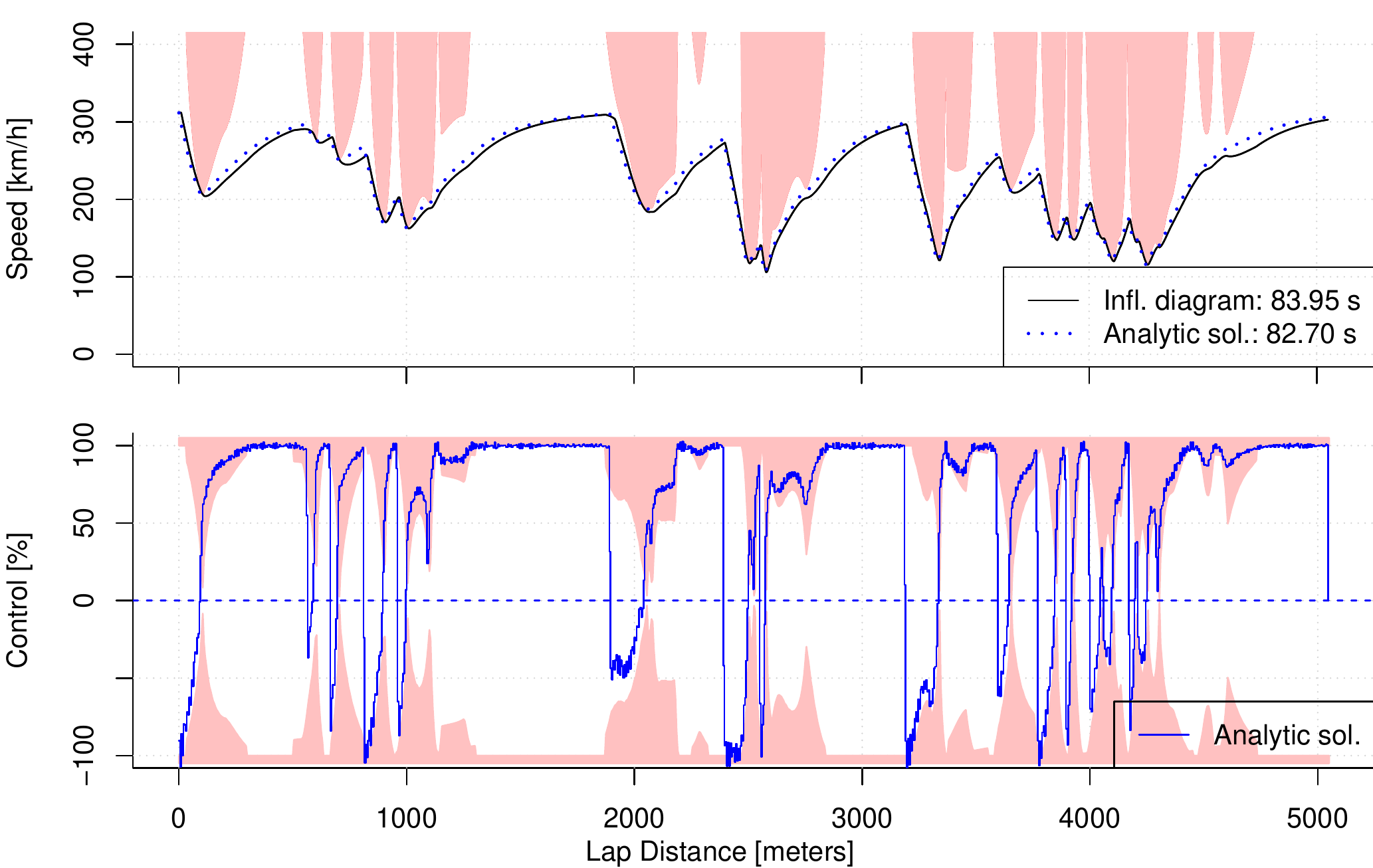}\\ 
\caption{A comparison of speed profiles: the influence diagram and analytic solution~\citep{velenis-tsiotras-2008}.
The shaded areas are forbidden by the speed and control constraints, respectively.
\label{fig-speed-profiles-2}}
\end{center}
\end{figure*}

The analytic solution 
was presented in~\citep{velenis-tsiotras-2008}. 
It is plotted in Figure~\ref{fig-speed-profiles-2} by a dotted line 
together with the influence diagram solution. 
The solutions are quite similar but there are some differences. 
Apparently, the analytical solution does not fully comply 
with the acceleration constraints. 
This causes differences in the speed profiles, otherwise they would be equivalent.
We were not able to explain this observation.
In the lower part of Figure~\ref{fig-speed-profiles-2} we present
the control profile of the analytic solution reconstructed 
from the speed profile of the analytic solution\footnote{The little oscillations
are caused by imprecision of the analytic speed profile taken from~\citep{velenis-tsiotras-2008}.}.

\section{RELATED WORK}\label{sec-related-work}

The speed profile optimization problem can be also specified using Markov decision processes (MDPs)~\citep{puterman-1994}
with a finite horizon, a non-stationary policy and a non-linear stationary reward function.
The solution of such an MDP can be found by the approach presented in this paper
since solution methods of both approaches are based on dynamic programming. 
In~\citep{velenis-tsiotras-2008} the problem was solved analytically 
by methods of the continuous time control~\citep{bertsekas-2000} using the Pontryagin's maximum principle.
We compared the influence diagram solution with the analytic solution
in Section~\ref{sec-experiments-analytic} -- the solutions were similar.
However, analytical solutions for considered extended versions of the problem 
with an advanced optimization function and additional constraints are not known 
and numerical methods have to be used.

\section{CONCLUSIONS AND FUTURE WORK\label{sec-conlusions}}

We proposed an application of influence diagrams to 
speed profile optimization and tested it in a real-life scenario.
We summarize what we have achieved and what we have learned:
\begin{itemize}
\item We were able to find optimal solutions efficiently.
\item We verified the solutions are in accordance
with the analytical solution of the considered problem.
\item An important advantage of influence diagrams is that once the policy is computed, 
it can be immediately used to update the optimal speed profile under modified circumstances. 
For example, if the driver has to slow down because of an unexpected traffic situation, 
the policy immediately provides the best new control value and 
the speed profile is specified by following the precomputed 
optimal policies. 
\item Influence diagrams are especially handy in more complex real-life scenarios 
where the analytic solution is unknown.
\item In applications, different optimality criteria come into play. 
We can do the computations efficiently as long as they decompose
additively along the path segments.
\end{itemize}
In future we plan to optimize speed profiles using influence diagrams 
with continuous variables.
Inspired by the work of~\citep{kveton-et-al-2006} on MDPs 
we plan to study inference in influence diagrams based on mixtures of beta distributions.
Other possibilities to be considered are approximations by mixtures of polynomials~\citep{shenoy-west-2011,li-shenoy-2012}
or by mixtures of truncated exponentials~\citep{cobb-shenoy-2008,moral-et-al-2001}.

Currently, we are applying influence diagrams to a more complex scenario with a complex utility function, 
for which no analytical solution is known. Methods of the control theory, if applied to this scenario, 
thus need to rely on approximate numerical methods.

\subsection*{Acknowledgements}

This work was supported by the Czech Science Foundation through project 13--20012S.

\clearpage

\bibliographystyle{apalike}
\bibliography{bibliography}

\end{document}